\documentclass[conference, letterpaper, final]{IEEEtran}
\IEEEoverridecommandlockouts
\usepackage{cite}
\usepackage{amsmath,amssymb,amsfonts}
\usepackage{algorithmic}
\usepackage{graphicx}
\usepackage{textcomp}
\usepackage{xcolor}
\usepackage{url}
\usepackage{booktabs}
\usepackage{xfp}
\usepackage{siunitx}
\usepackage{multirow}
\usepackage[]{todonotes}
\usepackage{xspace}
\usepackage[verbose]{placeins}
\usepackage{graphicx} 
\usepackage{caption}
\usepackage{subcaption}
\usepackage[absolute,showboxes]{textpos}
\newcommand{\sysname}{base\xspace}%
\newcommand{\rpl}{residual\xspace}%
\newcommand{\baseline}{standard\xspace}%
\newcommand{\Rpl}{Residual\xspace}%
\newcommand{\Baseline}{Standard\xspace}%
\newcommand{\ft}{F1TENTH\xspace}%
\usepackage[nogroupskip,acronyms,nopostdot,style=super,nonumberlist,toc]{glossaries}
\newacronym{ai}{AI}{artificial intelligence}%
\newacronym{drl}{DRL}{deep reinforcement learning}
\newacronym{dl}{DL}{deep learning}
\newacronym{ml}{ML}{machine learning}
\newacronym{rl}{RL}{reinforcement learning}
\newacronym{ad}{AD}{autonomous driving}
\newacronym{av}{AV}{autonomous vehicle}
\newacronym{dnn}{DNN}{deep neural network}
\newacronym{ann}{ANN}{artificial neural network}
\newacronym{nn}{NN}{neural network}
\newacronym{dqn}{DQN}{deep Q-network}
\newacronym{rnn}{RNN}{recurrent neural network}
\newacronym{rdqn}{RDQN}{recurrent deep Q-network}
\newacronym{ddqn}{DDQN}{double deep Q-network}
\newacronym{marl}{MARL}{multi-agent reinforcement learning}
\newacronym{dmarl}{DMARL}{deep multi-agent reinforcement learning}
\newacronym{mdp}{MDP}{Markov decision process}
\newacronym{mlp}{MLP}{multilayer perceptron}
\newacronym{mpc}{MPC}{model predictive control}
\newacronym{its}{ITS}{intelligent transportation systems}
\newacronym{ttc}{TTC}{time-to-collision}
\newacronym{ddpg}{DDPG}{deep deterministic policy gradient}
\newacronym{vae}{VAE}{variational auto-encoder}
\newacronym{mas}{MAS}{multi-agent system}
\newacronym{mal}{MAL}{multi-agent learning}
\newacronym{per}{PER}{prioritized experience replay}
\newacronym{a2c}{A2C}{advantage actor critic}
\newacronym{sg}{SG}{stochastic game}
\newacronym{mg}{MG}{Markov game}
\newacronym{pomdp}{POMDP}{partially observable Markov decision process}
\newacronym{pomg}{POMG}{partially observable Markov game}
\newacronym{dpomdp}{dec-POMDP}{decentralized partially observable Markov decision process}
\newacronym{nrmse}{NRMSE}{normalized root-mean-square error}
\newacronym{ppo}{PPO}{proximal policy optimization}
\newacronym{gae}{GAE}{generalized advantage estimate}
\newacronym{rpl}{RPL}{residual policy learning}
\def\BibTeX{{\rm B\kern-.05em{\sc i\kern-.025em b}\kern-.08em
    T\kern-.1667em\lower.7ex\hbox{E}\kern-.125emX}}

\begin{document}
\title{Residual Policy Learning for Vehicle Control\\of Autonomous Racing Cars}
\author{Raphael Trumpp, Denis Hoornaert, and Marco Caccamo
\thanks{R. Trumpp, D. Hoornaert, and M. Caccamo are with the TUM School of Engineering and Design, Technical University of Munich, Germany {\tt\footnotesize \{raphael.trumpp, denis.hoornaert, mcaccamo\}@tum.de}.}%
}

\maketitle
\begin{abstract}
The development of vehicle controllers for autonomous racing is challenging because racing cars operate at their physical driving limit.
Prompted by the demand for improved performance, autonomous racing research has seen the proliferation of machine learning-based controllers.
While these approaches show competitive performance, their practical applicability is often limited.
Residual policy learning promises to mitigate this drawback by combining classical controllers with learned residual controllers.
The critical advantage of residual controllers is their high adaptability parallel to the classical controller's stable behavior.
We propose a residual vehicle controller for autonomous racing cars that learns to amend a classical controller for the path-following of racing lines.
In an extensive study, performance gains of our approach are evaluated for a simulated car of the \ft autonomous racing series.
The evaluation for twelve replicated real-world racetracks shows that the residual controller reduces lap times by an average of 4.55~\% compared to a classical controller and even enables lap time gains on unknown racetracks.
\end{abstract}


\section{Introduction}
\label{sec:introduction}
Autonomous racing series, such as the F1TENTH~\cite{okelly2020f1tenth} and Indy Autonomous Challenge, are at the frontier of autonomous racing development. In against-the-clock racing, the computed racing lines push the car's physics to its limits, requiring highly fine-tuned controllers to deliver the expected performance. On the one hand, classical controllers, e.g., PID control, only achieve adequate performance with tedious racetrack-dependent parameter tuning. On the other hand, modern model-based controllers like model-predictive control require accurate vehicle models and have high computational requirements. The advancing field of \gls*{drl} promises to overcome these limitations by learning model-free vehicle controllers through trial-and-error interaction and accumulating a reward signal. It has been shown that end-to-end approaches using \gls*{drl} \cite{fuchs2021super} on top of a perception module learn competitive racing behavior. However, the practical application of these methods is challenging due to the sim2real gap and ample training time.

The field of \gls*{rpl}~\cite{johannink2019residual} overcomes these hurdles by combining classical controllers with a \gls{drl}-based approach to learn residual actions that correct and improve the performance of the base controller. In the context of autonomous racing, \gls*{rpl} promises to combine the \textit{reliability} of classical control approaches with the \textit{adaptability} of learning-based controllers. The residual controller can learn the control action under various optimization targets, e.g., energy consumption or vehicle stability, alongside the ability to cope with variations in the environment in real-time by observing the current vehicle state and its environment. Opposed to classical \gls*{drl}, \gls*{rpl} has proven to bootstrap the learning process leading to reduced training time with high real-world performance~\cite{zeng2020tossingbot}.

Originally motivated by robotic applications \cite{johannink2019residual, zeng2020tossingbot}, \gls*{rpl} has recently gained attention in the automotive community. In~\cite{kerbel2022residual}, the authors leverage \gls*{rpl} to improve traditional power-train control policies for better fuel consumption. \Gls*{rpl} has also been applied for suspension control to benefit passenger comfort~\cite{hynes2020optimizing}. Closest to our work is the use of \gls*{rpl} for autonomous racing in \cite{zhang2022residual}. The authors propose the ResCar model that combines a modified artificial potential field controller with \gls*{rpl}. Evaluated for simulated F1TENTH cars, the performance is demonstrated in five replicated real-world racetracks showing reduced demand for training data compared to the DreamerV2-based approach in \cite{brunnbauer2022latent}. However, ResCar is trained per racetrack, leading to tedious retraining as the generalization capability of the approach is only discussed for limited scenarios. Similarly, end-to-end \gls*{drl} approaches \cite{fuchs2021super} offer low generalization capability while requiring higher computational power. In contrast, we show that our method generalizes to unknown racetracks directly. Augmenting model-predictive control for trajectory optimization with \gls{drl} \cite{bellegarda2020online} is a promising direction with high generalization potential; our work proves that high performance can also be achieved with a fully model-free pipeline and only local observation of a planned trajectory.

We propose the use of \gls*{rpl} to synthesize a vehicle controller for autonomous racing cars consisting of two modules: The first module acts as a \textit{\sysname} controller that has local access to planned waypoints to follow an optimized racing line; this racing line may be obtained by map data in advance or through online observation of the racing car's environment. The second module, the \textit{residual} controller, uses \textit{local} observations to learn a residual action to amend the \sysname controller's action -- the \rpl controller accounts for local optimization potential due to non-optimal tracking performance of the \sysname controller, or a non-optimal racing line, to optimize for an overall improved performance. This adaptability can be crucial for achieving the best lap times in competitive racing, as our simulated F1TENTH racing car shows.

Our main contributions can be summarized as follows:
\begin{itemize}
    \item Design of a novel residual controller combining \gls*{rpl} with path-following of a planned racing line in the \ft-gym simulator \cite{okelly2020f1tenth} for twelve real-world racetracks.
    \item The influence of \gls*{rpl} for improving vehicle handling is analyzed in a study on vehicle dynamics and lap time.
    \item We demonstrate the generalization capability of the proposed approach by evaluating its behavior in unknown racetracks and randomized starting positions. This study proves that the training environments are not just memorized; instead, a deep scene understanding is learned.
    \item We reason about how the proposed residual controller improves lap times by an average of 4.55~\% compared to the base controller. 
    \item The code is publicly available\footnote{\url{https://github.com/raphajaner/rpl4f110}.}.
\end{itemize}

\section{Background}
\label{sec:background}
In the following, the used vehicle dynamics model and vehicle controller for path-following are introduced along with the theoretical description of \gls*{rpl} using \gls*{drl} algorithms.

\subsection{Vehicle Dynamics}
\label{subsec:vec_dym}
The \ft-gym simulator's vehicle dynamics are derived from the CommonRoad framework~\cite{althoff2017commonroad} using a kinematic single-track model with tire slip; the vertical load of all four wheels, their individual spin and slipe, and non-linear tire dynamics are not represented in this model. The vehicle model's control input is defined as $a_t=[\dot\delta_{t}, \dot v_t]$ with $\dot\delta_{t}$ as the steering velocity of the steering angle $\delta_t$, and $\dot v_t$ as the acceleration of the vehicle at time $t$.

\subsection{Trajectory Generation}
When global map knowledge of the racetrack to drive is available, trajectory optimization techniques can be used to derive a (near) optimal racing line consisting of waypoints $w=\{x_{d}, y_{d}, \phi_{d}, v_{d}\}$ that define a global desired position $[x_{d}, y_{d}]$ and orientation $\phi_{d}$ together with a planned velocity profile $v_{d}$. The set of waypoints $W=\{w_1, w_2, \dots\}$ is obtained by minimization of a loss function, e.g., the minimum curvature of the racing line as discussed in~\cite{heilmeier2019minimum}.

\subsection{Vehicle Controller for Path-Following}
\label{subseq:pure_pursuit}
The task in path-following is to follow a planned trajectory with minimal error, i.e., the positional and velocity tracking error must be minimized. While a PID controller is commonly used to track the velocity profile, i.e., a vehicle acceleration $\dot v_t$ is calculated w.r.t. the velocity error, the derivation of a well-performing control law for the steering angle $\delta_t$ is more involved. The pure pursuit controller~\cite{coulter1992implementation} calculates $\delta_t$ based on a geometric definition of the reference trajectory. Given a set of planned waypoints $W$, the current lookahead point $w_t\in{W}$ at time $t$ is selected in a fixed lookahead distance $l_d$ relative to the vehicle position. Using the geometric center of the circle defined by the vehicle's rear axis, its length $l$, and the waypoint $w_t$, the steering angle is calculated by
\begin{equation}
    \delta_t = \arctan\bigg(\frac{2l\sin\alpha}{l_d}\bigg),
\end{equation}
with $\alpha$ as the angle between the line connecting the vehicle's rear with the lookahead point and the vehicle's orientation.

\subsection{Residual Policy Learning}
In model-free \gls*{drl}, the behavior of a stochastic agent is defined by a policy $\pi(a_t | s_t)$ that maps the observed state $s_t \in \mathcal{S}$ to a probabilistic action space $\mathcal{P}(\mathcal{A})$ of (continuous) actions $a_t \in \mathcal{A}$. The state transition $\mathcal{T}$ is a mapping $\mathcal{T}: \mathcal{S} \times \mathcal{A} \rightarrow \mathcal{P}(\mathcal{S})$ defining the transition probability from states to new states when an action $a_t$ is taken. Such transitions are evaluated by the reward function $\mathcal{R}$ with a scalar value $r_t$. Combined with the discount factor $\gamma$, a \gls*{mdp} with tuple  $(\mathcal{S}, \mathcal{A}, \mathcal{T}, \mathcal{R}, \gamma)$ is described. 

For \gls*{rpl}, the policy $\pi_{\text{RB}}: \mathcal{S} \to \mathcal{A}$ is defined with
\begin{equation}
\label{eqn:residual_def}
    \pi_{\text{RB}}(s_t) = a_{\text{R}, t} |_{a_{\text{R}, t} \sim \pi_{\text{R}}(\cdot | s_t)} + \mu_{\text{B}}(s_t),
\end{equation}
combining the deterministic \textit{base} policy $\mu_{\text{B}}(s_t): \mathcal{S} \to \mathcal{A}$ and corresponding action $a_{\text{B},t}$ with a \textit{learned} stochastic\footnote{For deterministic policies, (\ref{eqn:residual_def}) can be rewritten to $\pi_{\text{RB}}(s_t) = \pi_{\text{R}}(s_t) + \mu_{\text{B}}(s_t)$.} residual policy $\pi_{\text{R}}: \mathcal{S} \rightarrow \mathcal{P}(\mathcal{A})$. For $\mu_{\text{B}}(s_t)$, any deterministic control algorithm can be used while $\pi_{\text{R}}(a_t | s_t)$ is typically learned by \gls{drl}. Because $\pi_{\text{R}}(a_t | s_t)$ is defined as a probability distribution, actions must be sampled with $a_{\text{R},t} \sim \pi_{\text{R}}(\cdot | s_t)$. Note that for the environment to be Markovian, the base action $a_{\text{B},t}$ must be part of or be observable from $s_t$.

\subsection{Deep Reinforcement Learning Algorithms}
\Gls*{drl} combine the use of \glspl*{nn} with an iterative learning approach. Typically, a value $V(s_t)$ or Q function $Q(a_t, s_t)$ is learned through interaction with an environment and observable transition tuples $e_t=\{s_t, a_t, r_{t},s_{t+1}\}$ along a trajectory $\tau=\{e_0, e_1, ..., e_T\}$ of $T$ fixed-sized steps. 

The \textit{on}-policy algorithm \gls*{ppo}~\cite{schulman2017proximal} learns a stochastic policy $\pi_{\theta_k}$, parameterized with weights $\theta_k$ at iteration $k$, by updating the policy according to 
\begin{equation}
\label{eqn:ppo_update}
\theta_{k+1}=\arg \max _\theta \underset{s_t, a_t \sim \pi_{\theta_k}}{\mathrm{E}}\left[L\left(s_t, a_t, \theta_k, \theta\right)\right]
\end{equation}
with the objective function
\begin{align}
    L\left(s_t, a_t, \theta_k, \theta\right) = \min \biggl( &\frac{\pi_\theta(a_t \mid s_t)}{\pi_{\theta_k}(a_t \mid s_t)} A^{\pi_{\theta_k}}(s_t, a_t), \nonumber \\ 
     & g\left(\epsilon, A^{\pi_{\theta_k}}(s_t, a_t)\right) \biggr) 
\end{align}
where $g(\epsilon, A)$ is defined as
\begin{equation}
\label{eqn:g_eps}
g(\epsilon, A)= \begin{cases}(1+\epsilon) A & A \geq 0 \\ (1-\epsilon) A & A<0 .\end{cases},
\end{equation}
with the advantage function $A=A^{\pi_{\theta_k}}(s_t, a_t)$. In (\ref{eqn:g_eps}), $\epsilon$ is a hyperparameter that implies that a gradient update step has limited incentive for the new policy to be far away in probability space from the previous one through clipping of the function, i.e., policy changes are regularized. This version of \gls*{ppo}, often called \gls*{ppo}-Clip, requires calculating the advantage function $A$. Typically, the \gls*{gae}~\cite{schulman2015high} approach is used which requires the training of a value function $V_{\phi_k}(s_t)$ with weights $\phi_k$ in parallel to the optimization of the policy $\pi_{\theta_k}$. In practice, the expectation in (\ref{eqn:ppo_update}) is estimated by batches of $B$ recorded $T$-step long trajectories $\mathcal{D}_k=\{\tau_0, \tau_2, \dots, \tau_{B-1}\}_{k}$ that must have been realized by using the current policy $ \pi_{\theta_k}$. The value function $V_{\phi_k}$ is updated by regression of
\begin{equation}
    \phi_{k+1}=\arg \min _\phi \frac{1}{\left|\mathcal{D}_k\right| T} \sum_{\tau \in D_k} \sum_{t=0}^T\left(V_{\phi_k}\left(s_t\right)-\hat{R}_t\right)^2,
\end{equation}
with the reward-to-go $\hat{R}_t=\sum_{t'=t}^{T}r_t$.
\section{System Design}
\label{sec:problem_statement}
\begin{figure}[!t]
    \centering
    \includegraphics[width=0.45\textwidth]{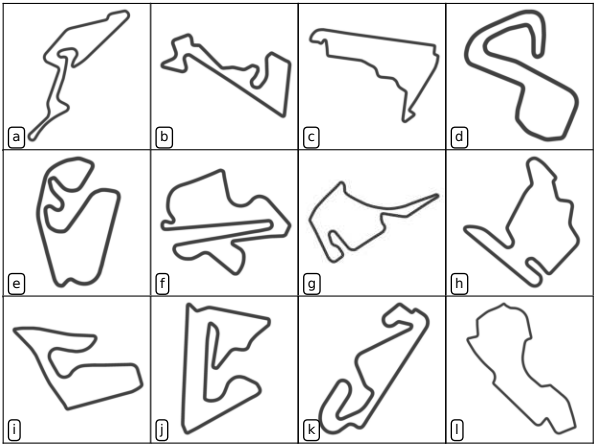}
    \caption{Racetracks for training and testing: (a)~N\"urburgring, (b)~Moscow Raceway, (c)~Mexico City, (d)~Brands Hatch, (e)~Sao Paulo, (f)~Sepang, (g)~Hockenheim, (h)~Budapest, (i)~Spielberg, (j)~Sakhir, (k)~Catalunya, and (l)~Melbourne.}
    \label{fig:raectracks-collection}
\end{figure}%
This work proposes an \gls*{rpl}-based controller for vehicle control of autonomous racing cars. The approach is developed in a simulator for \ft cars.
In the following, the simulation environment and the components of the developed vehicle controller are presented.

\subsection{Learning Environment}
The \ft-gym environment \cite{okelly2020f1tenth} simulates vehicle dynamics using the single-track model (including slip) as presented in Section \ref{subsec:vec_dym}. The simulator allows the use of arbitrary map data; we use the replication of twelve real-world racetracks from the F1TENTH-racetracks~\cite{betz2022autonomous} repository\footnote{\url{https://github.com/f1tenth/f1tenth_racetracks}, commit: \texttt{b95c4ef}.} adjusted to the \ft scale. As it can be seen in their visualization in Fig. \ref{fig:raectracks-collection}, these racetracks offer various track designs with challenging high- and low-speed sections.

These real-world F1TENTH-racetracks offer an associated racing line for each racetrack in the form of a set of waypoints that include their global positions, an optimized velocity profile, and curvature information. According to the authors of the repository, the racing lines have been obtained via minimum curvature trajectory planning such as described in ~\cite{heilmeier2019minimum}. We use realistic vehicle parameters as measured from our real \ft car; see Table \ref{tab:car_model_parameters} for a full list of the used parameters. The simulator runs with a synchronous simulation and control frequency of 100~Hz. As implemented in the simulator, a low-level controller is used that sets the model input $\left[\dot\delta_t, \dot v_t\right]$ given a high-level control action $a_t=\left[\delta_t, v_t\right]$.
\begin{table}[!t]
    \centering
    \caption{Vehicle model parameters for a \ft RC-car.}
    \label{tab:car_model_parameters}
    \begin{tabular}{lrll}
    \toprule
    Parameter                 & Value   & Unit                          & Description                         \\
    \midrule
    $h$                       & 0.074   & m                             & Height                              \\
    $w$                       & 0.27    &  m                            & Width                               \\
    $l$                       & 0.51    &  m                            & Length                              \\
    $l_f$                     & 0.15875 & m                             & Length from COG to front            \\
    $l_r$                     & 0.17145 & m                             &    Length from COG to rear          \\
    $m$                       & 3.47    & kg                            & Total mass                          \\
    $I$                       & 0.04712 & kg$\cdot$m\textsuperscript{2} & Vehicle inertia                     \\
    $\delta^{\text{max}}$     & 0.4189  & rad                           & Steering angle bounds               \\
    $\dot\delta^{\text{max}}$ & 3.2     & rad/s                         &  Steering velocity bounds           \\
    $v^{\text{switch}}$       & 7.319   &  m/s                          & Dynamic bound for full acceleration \\
    $v^\text{max}$            & 8.0     &  m/s                          & Maximal velocity                    \\
    $\dot v^\text{max}$       & 7.51    &  m/s\textsuperscript{2}       & Maximal acceleration                \\
    $C_{S,f}$                 & 4.718   & N/m                           & Tire stiffness front                \\
    $C_{S,r}$                 & 5.4562  & N/m                           & Tire stiffness rear                 \\
    $\mu$                     & 0.8     & -                             & Friction coefficient (estimated)    \\
    \bottomrule
\end{tabular}
\end{table}
\subsection{Residual Controller}
\begin{figure}[!t]
    \centering
    \includegraphics[width=0.4\textwidth]{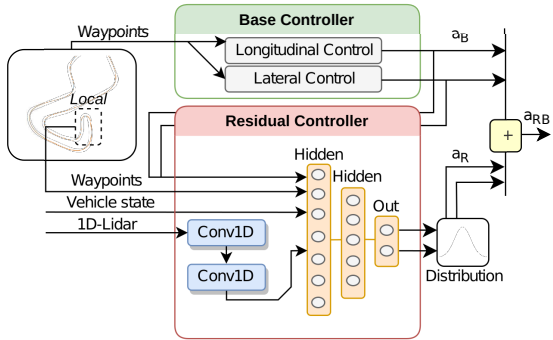}
    \caption{Proposed control architecture consisting of a \sysname controller that outputs control action $a_{\text{B}}$ which is added to the residual action $a_{\text{R}}$ to form the combined action $a_{\text{RB}}$. The action $a_{\text{R}}$ of the residual controller is sampled from a distribution whose parameters are defined by a \gls*{nn}.}
    \label{fig:network}
\end{figure}

We proposed a residual controller architecture that consists of two sequential control modules:
\begin{enumerate}
    \item The \emph{\sysname} controller consisting of a classical controller outputting a deterministic control action $a_{\text{B},t}$.
    \item The \emph{\rpl} controller that learns a control action $a_{\text{R},t}$ by \gls*{rpl} and interaction with the environment.
\end{enumerate}
As shown in Fig. \ref{fig:network}, our presented controller obtains the control action $a_{\text{RB}, t}$ by combining both control modules
\begin{equation}
    a_{\text{RB},t} = a_{\text{R},t} + a_{\text{B},t},
\end{equation}
with $a_{\text{RB},t}=[v_t, \delta_t]$ as the high-level control input to the simulator's low-level controller. 

The \sysname module's action $a_{\text{B},t}$ is derived from a pure pursuit controller for lateral control; see Section \ref{subseq:pure_pursuit}. The steering angle $\delta_{t}$ is calculated by minimizing the tracking error of the nearest waypoint of the planned racing line along with the velocity taken from the planned velocity profile of this waypoint; a lookahead distance of $0.82$~m is used. 

The \rpl controller is implemented by a \gls*{ppo} agent which observes only local information
\begin{align}
        s_t = \big[&L_t,w_{\text{rel},t:}, v_{x,t}, v_{y,t}, \dot v_{x,t}, \dot v_{y,t}, \nonumber \\
             &\psi_{t}, \dot\psi_{t}, \beta_{t}, a_{\text{B},t}, a_{\text{RB},t-1}\big],
\end{align}
with $L_t\in\mathcal{R}$ as a 1D-lidar representation with 1080 points spread along a 270$^{\circ}$ field-of-view measuring the vehicle's distance to other objects, e.g., the wall defining the racetrack. A limited set of future waypoints $w_{\text{rel}, t:}$ are given with relative coordinates to the vehicle's current position. This information is limited to local observation; only the waypoints ahead in 30~m distance are considered. The vehicle state with longitudinal and lateral velocity and acceleration, respectively, is concatenated to the yaw angle $\psi$, yaw rate $\dot\psi$, and the slip angle $\beta$. Additionally, the agent observes the action planned by the \sysname controller $a_{\text{B},t}$ and the previous action that was applied $a_{\text{RB},t-1}$. We stack the last three frames to introduce a limited memory to the agent; only the recent observation of $L_t$ and $w_{\text{rel},t:}$ at $t$ are used to reduce the dimension of the state. The reward function used for learning is defined as
\begin{equation}
    r_{t+1} = \tau_{1} \cdot v_{x,t} + \tau_{2} \cdot v_{y,t}^2 + \rho \cdot \left\{\begin{array}{ll} 1, & \text{if collision}_{t}=\text{True} \\ 0, & \text{otherwise}\end{array}\right.\\,
\end{equation}
with $\tau_{1}=0.003$ to encourage the agent to optimize for higher velocity while not increasing the lateral velocity, and therefore instability, too much due to $\tau_{2}=-0.003$. Moreover, $\rho=-50$ is set to prevent collisions.
\subsection{Network Design}
In \gls*{ppo}, the policy $\pi_{\theta}$ is defined as a \gls{nn} with weights $\theta$ along with a value network $V_{\phi}$ and its parameters $\phi$. As shown in Fig. \ref{fig:network}, our proposed architecture uses a perception module in form of two consecutive layers of 1D-convolution with 16 and 32 filters, respectively, with ReLu activation and average pooling. The perception module learns an embedding of the 1D-lidar data $L_t$; its weights are shared between the policy and value network. In the policy and value network, respectively, the \glspl{mlp} consists of two hidden linear layers with $\{400,300\}$ neurons and ReLu activation and an output layer of fitting output dimension. The policy network learns the parameters of a TanhNormal distribution, i.e., a Gaussian distribution which is $\tanh$ transformed for projection to a bounded space $[-1, 1]$, with state-depended mean $\mu(s_t)$ but state-independent variance $\sigma^2$. 

\subsection{Training Procedure}
\label{sec:training-procedure}
We use a customized version of \gls*{ppo} as implemented in \cite{huang2022cleanrl}. The residual controller is trained for $1e7$ episodes which correspond to approximately 2000 full race laps; the high-control frequency of the simulator requires training with long trajectories of 2048 steps. Training trajectories are collected from 36 environments running in parallel. Observations and rewards are normalized by a running statistics calculation. While we set $\gamma=0.998$, a batch size of 128, and stop epoch updates early for steps larger than a KL-divergence of 0.01, other used hyperparameters are unchanged from \cite{huang2022cleanrl}. A scaling factor of $[0.05, 1]^{\top}$ is multiplied with the policy output to bound the residual $a_{\text{R},t}$ for the steering angle and the vehicle velocity by $\pm~0.05$ and $\pm~1.0$, respectively. We use this setting for safety reasons to prevent an overly large influence of the residual controller on the base controller's behavior.

To evaluate the generalization capability of the proposed residual controller, training is only done on \textit{nine} training racetracks (a-i) while the \textit{three} test racetracks (j-l) are used for validation and are not seen during training. The environment resets after two full rounds on a racetrack are completed. Additional variability is introduced by randomly selecting a waypoint on the racing line every time the environment resets with a new starting position. While this makes the driving task more challenging, it is crucial for a meaningful evaluation of the controller's generalization capabilities.

\begin{figure*}[!thb]
    \begin{minipage}[!b]{0.65\textwidth}
            \begin{subfigure}{\textwidth}
                \centering
                \includegraphics[width=0.9\textwidth]{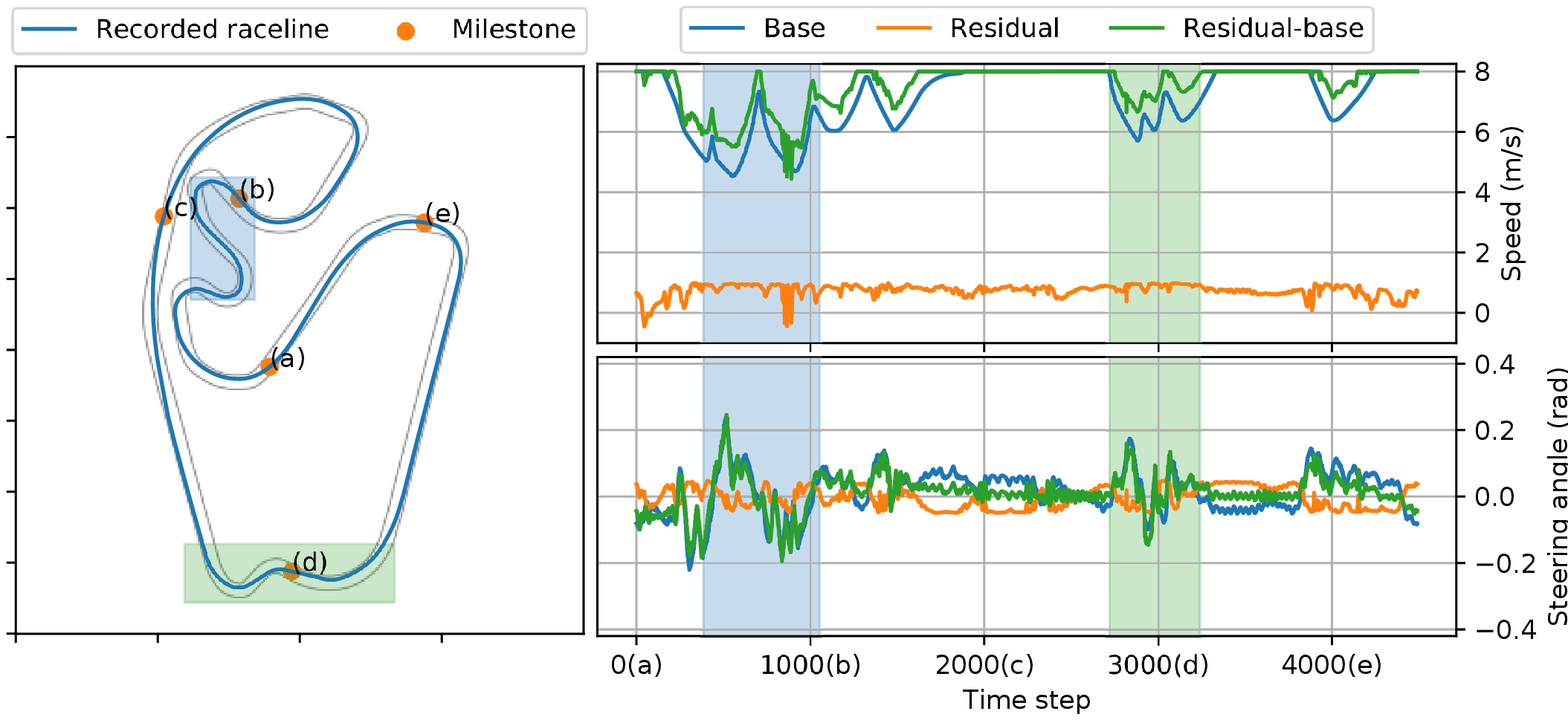}
                \caption{Action profile for the Sao Paulo racetrack.}
                \label{fig:trajectory-recording-saopaulo}
            \end{subfigure}%
            \hfill%
            \begin{subfigure}{\textwidth}
                \centering
                \includegraphics[width=0.9\textwidth]{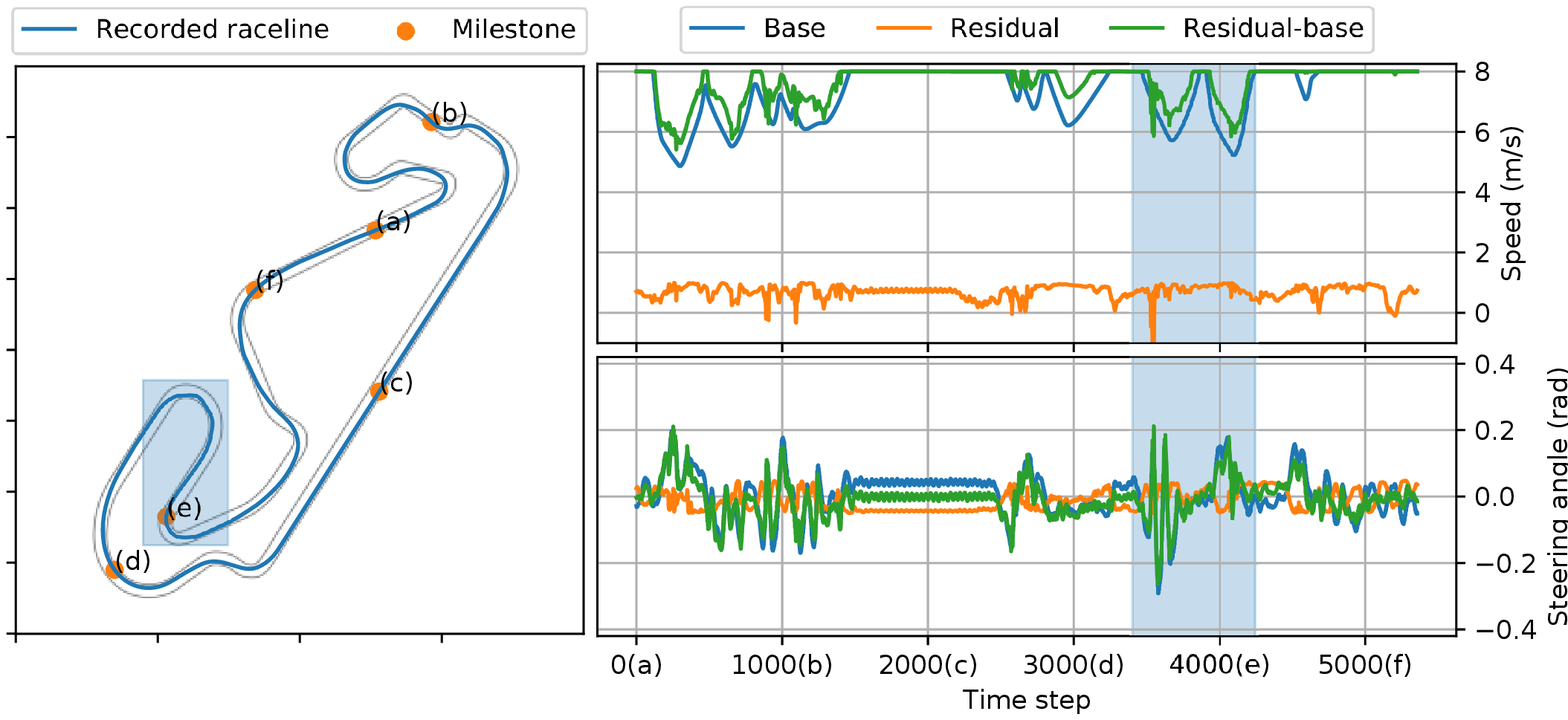}
                \caption{Action profile for the Catalunya racetrack.}
                \label{fig:trajectory-recording-spielberg}
            \end{subfigure}%
            \caption{Overview of the residual controller's action profile split into base $a_{\text{B},t}$ (blue line), residual $a_{\text{R},t}$ (orange line), and residual-base $a_{\text{RB},t}=a_{\text{R},t}+a_{\text{B},t}$ (green line). Milestones (a) to (f) on the left inset correspond to the profiles (right insets) time steps (x-axis ticks). Green and blue areas provide a more precise overview of the vehicle behavior in the corresponding highlighted regions of the racetrack.}
            \label{fig:trajectory-recording}
    \end{minipage}%
    \hfill
    \begin{minipage}[!b]{0.33\textwidth}
        \centering
        \includegraphics[width=\textwidth]{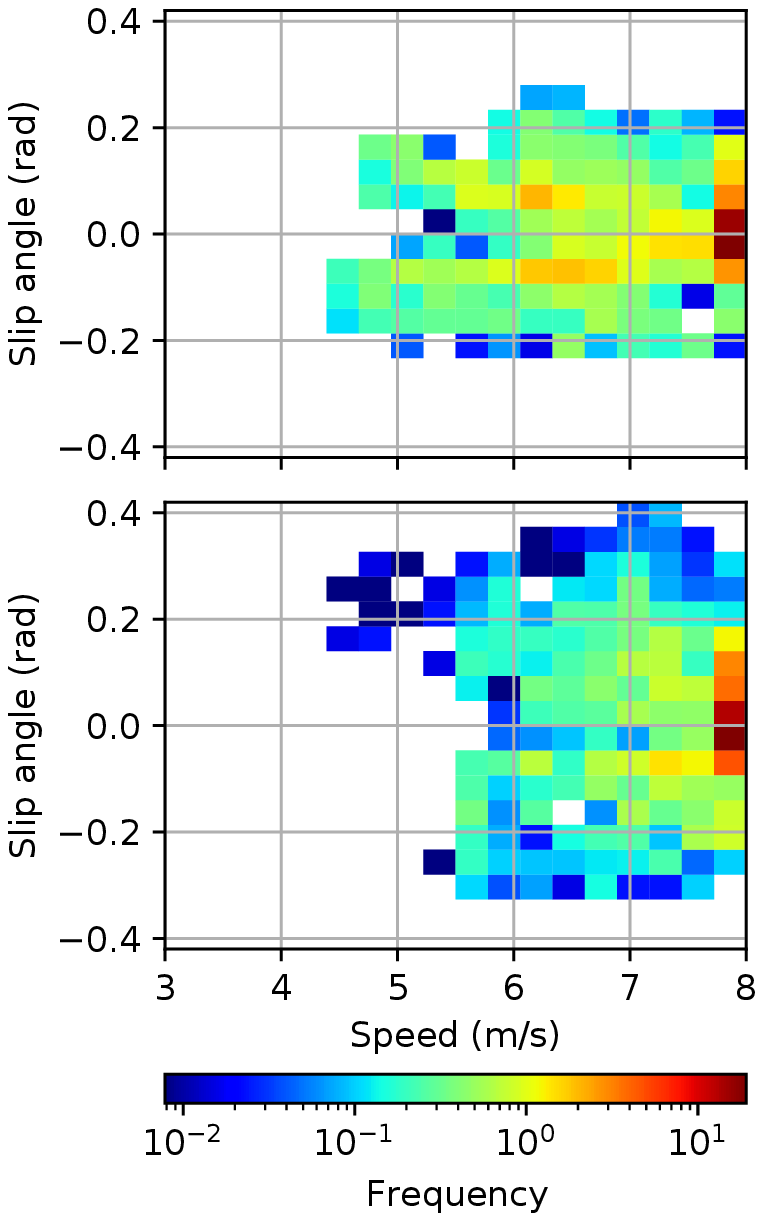}
        \caption{Frequency of vehicle slip $\beta$ aggregated over single laps on the Sao Paulo and Catalunya racetracks. The \sysname controller (top inset) has a maximum $|\beta|$ of $0.27$~rad while the \rpl controller's higher velocity (bottom) comes at the cost of a higher slip angle up to $0.43$~rad.}
        \label{fig:slip-angle-cartography}
    \end{minipage}%
\end{figure*}
\section{Evaluation}
    \label{sec:evaluation}
    This section assesses the improvements linked to the use of our residual controller. For evaluation, the mean of the residual controller's stochastic policy $\pi_{\theta}$ is taken instead of sampling from the learned distribution. We also introduce the \textit{\baseline} controller as a baseline that uses the \sysname controller's action $a_{\text{B},t}$ but without the added residual controller for comparison. 
    
    First, a high-level comparison of the lap time for all available racetracks is provided in Section~\ref{sec:laptimes-results}. Reported lab times are the median results of two running start laps with random starting points over three independent full training runs, i.e., six lap time estimates overall, with changing seeds to estimate training stability. Secondly, Section~\ref{sec:rpl-behaviour} presents an analysis of trajectories for a sub-set of racetracks for the best of the three trained controllers. Finally, Section~\ref{sec:residual-impact} compares the stability of this controller with the baseline. 

    \subsection{Lap Times}
        \label{sec:laptimes-results}
        In this experiment, the \rpl controller's performance gains are quantified against the \baseline controller w.r.t. recorded lap times.
        Table~\ref{tab:lap-times-results} shows that lap times for the training set are directly improved by adding the \rpl component. The relative improvements range from 2.07~\% to 7.06~\% with an overall average improvement of 4.55~\%. While improvements are expected for the training racetracks, similar gains can also be observed when the \rpl controller is deployed on the test set of racetracks. Lap times on the Sakhir, Catalunya, and Melbourne racetracks are improved by 4.34~\%, 5.24~\%, and 2.44~\%, respectively. 
        This evaluation demonstrates the capability of our proposed approach to generalize to unknown racetracks with comparable performance to the results observed on the training racetracks.
        
        Looking closer at the characteristics of the racetracks and the achieved relative improvements, a correlation between the racetrack curvature and the relative improvements can be observed. Racetracks with high curvatures, such as Moscow, Sao Paulo, and Hockenheim, form the top-3 improvements, whereas low curvature racetracks like Brand-Hatch tail the ranking. A similar observation can be made for the test set, with the Catalunya racetrack presenting a higher curvature than Melbourne and, thus, a higher lap time improvement. 
        \begin{table}[!t]
            \centering
            \caption{Lap time results of the \emph{\baseline} and the \emph{\rpl} controllers for all racetracks as medians of three training runs.}
            \label{tab:lap-times-results}
            \begin{tabular}{cclrrrr}
    \toprule
    & \multicolumn{2}{c}{\textbf{Racetrack}} & \multicolumn{2}{c}{\textbf{Lap time}} & \multicolumn{2}{c}{\textbf{Improvement}} \\
    & \# & Name          & \Baseline & \Rpl & Diff.            & Rel.\\
    \midrule
    \multirow{8}{*}{\rotatebox[origin=c]{90}{Training}} 
    & a & N\"urburgring &     60.84 s &    58.07 s & 2.77 s & 4.55 \% \\
    & b & Moscow        &     46.75 s &    43.45 s & 3.30 s & 7.06 \% \\
    & c & Mexico City   &     49.12 s &    46.76 s & 2.36 s & 4.80 \% \\
    & d & Brands Hatch  &     45.92 s &    44.97 s & 0.95 s & 2.07 \% \\
    & e & Sao Paulo     &     47.92 s &    44.92 s & 3.00 s & 6.26 \% \\
    & f & Sepang        &     66.24 s &    63.18 s & 3.06 s & 4.62 \% \\
    & g & Hockenheim    &     49.96 s &    47.35 s & 2.61 s & 5.22 \% \\
    & h & Budapest      &     54.33 s &    51.67 s & 2.66 s & 4.90 \% \\
    & i & Spielberg     &     45.33 s &    43.93 s & 1.40 s & 3.09 \% \\
    \midrule
    \multirow{3}{*}{\rotatebox[origin=c]{90}{Test}} 
    & j & Sakhir        &     60.34 s &    57.72 s & 2.62 s & 4.34 \% \\
    & k & Catalunya     &     56.50 s &    53.54 s & 1.49 s & 5.24 \% \\
    & l & Melbourne     &     61.03 s &    59.54 s & 2.96 s & 2.44 \% \\
    \hline
    \hline
    \multicolumn{3}{r}{Overall averages} & 53.69 s & 50.85 s & 2.43 s &  4.55 \%\\
    \bottomrule
\end{tabular}
        \end{table}
        
    \subsection{\Rpl Controller Behaviour}
        \label{sec:rpl-behaviour}
        Given the hypothesis that the \rpl approach improves lap times when handling curves, we analyze the \rpl controller's behavior profile in action-state space for one racetrack each from the training and test set.
        
        Fig.~\ref{fig:trajectory-recording} displays the vehicle's recorded trajectory for the Sao Paulo and Catalunya racetracks w.r.t. the action profile for the \rpl controller, split into the action $a_{\text{B}, t}$ of the \textit{base} controller, the pure \textit{residual} action $a_{\text{R}, t}$ itself, and the combination of both as \textit{residual-base} $a_{\text{RB}, t}$. Overall, it can be seen that the \rpl controller (i) has a strong tendency to increase the velocity; the action oscillates close to $1~\frac{\text{m}}{\text{s}}$, and (ii) only decreases the velocity in rare occurrences.
        Note that velocities above $v^{\text{max}}=8~\frac{\text{m}}{\text{s}}$ are clipped due to the racing car's physical limits, which occurs when the racing line's planned velocity profile is already close to this limit. 
        
        For the Sao Paulo racetrack in Fig.~\ref{fig:trajectory-recording-saopaulo}, it can be seen that even in the curvy section in the blue area of interest, the \rpl controller increases the velocity set by the base controllers most times. Interestingly, the \rpl controller has learned to decelerate the car when approaching the curves' apex. This behavior yields more efficient steering behavior and a more aggressive racing line. Similar behavior is observed in the s-shaped section highlighted by the green area of interest, where the joint optimization of velocity and steering allows for taking dynamics effects into account. Furthermore, as can be seen from before milestone (c) and up to milestone (d), the \rpl controller has learned to comprehend the vehicle dynamics as it astutely counterbalances the base controller's steering to alter the trajectory enabling stable and high velocities.
        
        Evaluation of the Catalunya racetrack in Fig.~\ref{fig:trajectory-recording-spielberg} supports these findings.
        As highlighted in the blue area of interest, the \rpl controller reduces the velocity before the curve's apex. Analysis of the straight section between milestones (b) and (d) validates that the \rpl controller has learned to adequately counter-steer at high velocities.
        Observing the same characteristics on a racetrack issued from the test set further confirms that the \rpl agent is capable of generalization.
    
    \subsection{Vehicle Stability}
        \label{sec:residual-impact}
        Based on the analysis in Section~\ref{sec:rpl-behaviour}, the observed improvements in lap times can be credited to higher velocity when taking turns. Doing so incurs more constraints on the racing car, potentially leading to reduced control quality and a detrimental deterioration of the car's stability. To evaluate the latter, we monitor the slip angle $\beta$ of the kinematic single-track model with a linear relation to the tire forces. While a small slip angle is desirable w.r.t. stability, overall racing performance can only be improved by exploiting the car's dynamics adequately.
        Fig.~\ref{fig:slip-angle-cartography} reports an aggregation of the experienced slip angles when using the \rpl controller. Compared to the standard controller as a baseline, it is clear that the \rpl controller experiences larger slip angles in the range of $[-0.31, 0.43]$~rad compared to the \baseline controller's range of $[-0.23, 0.27]$~rad.
        This observation confirms the intuition described above that higher speeds and short lap times are achieved at the expense of the car's stability. However, the fact that the residual controller can exploit the stability limit to its advantage without causing collisions demonstrates the added value of the \rpl component and its capability to adapt from known to unknown racetracks.
\section{Conclusion}
    \label{sec:conclusion}
    We presented a new approach for vehicle control of autonomous racing cars using \gls*{rpl}. The proposed architecture consists of a \textit{base} module for path-following of a racing line and a \textit{residual} controller that learns a residual action from local observation to amend the base controller. Our study on vehicle behavior shows that the residual controller improves lap times by 4.55~\% in average over a set of twelve replicated real-world racetracks. These improvements can be credited to adequate vehicle control in high curvature sections enabled by the residual controller. Moreover, the zero-shot generalization capability of our approach is demonstrated on unknown racetracks. Additionally to an ablation study of the \rpl design, the envisioned future work will aim at evaluating the improvements of the residual controller for an extended set of scenarios. This includes the simulation of non-linear effects, e.g., drift, as well as a robustness analysis for noisy observations. Finally, we plan to explore the applicability of \gls*{rpl} to bridge the sim2real gap on real \ft cars.
\section*{Acknowledgement}
    \label{sec:acknowledgements}
    Marco Caccamo was supported by an Alexander von Humboldt Professorship endowed by the German Federal Ministry of Education and Research.

\bibliographystyle{IEEEtran}
\bibliography{references}

\end{document}